\documentclass[11pt]{article}
\usepackage{coling2020}
\usepackage{times}
\usepackage{url}
\usepackage{bbding}
\usepackage{latexsym}
\usepackage{graphicx}
\usepackage{stfloats}
\usepackage{amsthm,amsmath,amssymb}
\usepackage{mathrsfs}
\usepackage{tabularx}
\usepackage{color}
\usepackage{comment}
\usepackage{booktabs}
\usepackage{multirow}
\usepackage{adjustbox}
\usepackage{bm}
\usepackage[utf8]{inputenc}
\usepackage[normalem]{ulem}
\useunder{\uline}{\ul}{}

\colingfinalcopy 

\title{Exploit Multiple Reference Graphs for Semi-supervised Relation Extraction}

\author{Wanli Li \\
  School of Computer Science \\
  Wuhan University \\
  China \\
  {\tt wanli.li@whu.edu.cn} \\\And
  Tieyun Qian \\
  School of Computer Science \\
  Wuhan University \\
  China \\
  {\tt qty@whu.edu.cn}}

\date{}

\begin{document}
\maketitle
\begin{abstract}Manual annotation of the labeled data for relation extraction is time-consuming and labor-intensive. Semi-supervised methods can offer helping hands for this problem and have aroused great research interests. Existing work focuses on mapping the unlabeled samples to the classes to augment the labeled dataset.  However, it is hard to find an overall good mapping function, especially for the samples with complicated syntactic components in one sentence.

To tackle this limitation, we propose to build the connection between the unlabeled data and the labeled ones rather than directly mapping the  unlabeled samples to the classes. Specifically, we first use three kinds of  information to construct reference graphs, including entity reference, verb reference, and semantics reference. The goal is to semantically or lexically connect the unlabeled sample(s) to the labeled one(s). Then, we develop a Multiple Reference Graph  (MRefG) model to exploit the reference information for better recognizing high-quality unlabeled samples.
The effectiveness of our method is demonstrated by extensive comparison experiments with the state-of-the-art baselines on two public datasets.
\end{abstract}

\section{Introduction}
Relation Extraction (RE) aims to predict the relation between entities in the text. RE tasks can be categorized into sentence-level and document-level types, where two entities belong to one sentence and multiple sentences, respectively. In this paper, we focus on the sentence-level RE task. For example, given a sample ``\emph{People is moving back into downtown}'', there are  two entities \emph{`e1: People'} and \emph{`e2: downtown'}, and the goal is to predict the relation between these two entities, i.e., \emph{`entity:destination (e1, e2)'}. RE has aroused much research attention in recent years due to its broad applications in many machine learning and natural language processing tasks such as knowledge graph  construction \cite{DBLP:journals/vldb/SaRR0W0Z17:2}, biomedical knowledge discovery \cite{DBLP:conf/eacl/QuirkP17:1}, question answering \cite{DBLP:conf/acl/YuYHSXZ17:3}, and information retrieval \cite{DBLP:conf/sigir/KadryD17:4}.

Previous studies in RE mainly concentrate on designing  supervised learning \cite{DBLP:journals/jmlr/ZelenkoAR03:5,DBLP:conf/emnlp/ZengLLS17:6,DBLP:conf/acl/GuoZL19:35} or distance supervision \cite{DBLP:conf/tal/ReiplingerWK14:48,DBLP:conf/aaai/Ji0H017:49,jia-etal-2019-arnor:50} methods. Supervised learning requires a sufficient number of labeled data to train a good model. Hence it involves manual annotation and is labor-intensive. Distance supervision employs the knowledge bases (KB) to automatically annotate two entities co-occurring in one sentence with their KB relation. Distance supervision methods can alleviate the problem of manual labelling, but their main drawback lies in that KB relations may not be always applicable to the samples in RE tasks.

For developing general annotation method without the intervention of knowledge bases, semi-supervised learning ~\cite{Jones_Bootstrapping99,Agichtein_Snowball00} can offer helping hands. Most of existing studies along this line use rule-based approaches with bootstrapping framework~\cite{Batista_Bootstrapping15,Gupta_Bootstrapping18,DBLP:conf/iclr/WangQZ0YN0R20}. However, the rules still need to be pre-defined or manual annotating. A recent work \cite{DBLP:conf/www/Lin0QR19:13} presents a novel semi-supervised and rule-free approach DualRE to augment the training set by finding the high-quality samples from the unlabeled data. DualRE has achieved great successes in several public RE benchmarks.

While being powerful, the DualRE model suffers from one severe problem, i.e., the quality of augmented data highly depends on the mapping function $y=f(x)$, where $x$ is the feature vector of the instance in unlabeled data and $y$ is the relation of the entities in $x$. Note that $x$ is transformed into the feature vector using an encoder and the function $f()$ is learnt using the training corpus. That is to say, only if the sample $x_u$ in the unlabeled data is close to the sample $x_t$ in the training data in the latent vector space, can the function $f()$ assign the entities in  $x_u$ to the same relation $y$ of the entities in $x_t$. However, the position in which an instance is located is determined by all components in the instance, and thus it is easily affected by various modifiers. For example, the unlabeled sample ``19-year-old richest man \emph{Bill Gates} founded the great \emph{Microsoft} company in 1975'' is far away from the labeled sample ``\emph{Steve Jobs} founded \emph{Apple}''. Consequently, it is hard for the mapping function to assign the correct `\emph{org:founded\_by} (\emph{Bill Gates}, \emph{Microsoft})' relation to the unlabeled sample.

In order to tackle this limitation, we propose to build the connection between the unlabeled data and the labeled ones rather then directly mapping the  unlabeled data to the class labels. Our key observation is that the neighboring instances in the vector space should be semantically and/or lexically related. The semantic relation can be measured by the cosine or dot product similarity between two vectors, and the lexical relation should be  determined by the main components of the sentences. Since verbs play a vital role in grammar and entities are critical to the RE task, we choose the verb and entity as the main components to capture the lexical relations.

To model the above ideas, we first construct three types of reference graphs, including entity reference, verb reference, and semantics reference, to connect the unlabeled sample(s) to the labeled one(s) based on their semantic or lexical relations. We then develop a Multiple Reference Graph  (MRefG) model to exploit the reference information for better recognizing high-quality unlabeled samples, which will be used as the augmented data during the semi-supervised learning process. We conduct extensive experiments on two real world datasets. Results prove that our method outperforms the state-of-the-art supervised and semi-supervised baselines.

\section{Related Work}
\paragraph{Relation Extraction (RE)}
Early studies on RE mostly use traditional machine learning algorithms  \cite{zelenko-etal-2002-kernel:14,DBLP:journals/ml/BikelSW99:15,DBLP:conf/icml/McCallumFP00:16,DBLP:conf/icml/LaffertyMP01:17}. With the success of deep learning  techniques, many neural models are proposed for RE tasks \cite{socher-etal-2012-semantic:18,DBLP:conf/naacl/NguyenG15:19,DBLP:conf/acl/ZhouSTQLHX16:20,DBLP:conf/emnlp/ZengLC015:21}.

The aforementioned methods are all supervised based and require a large number of labeled data  to train an effective classifier. To address this problem, there are mainly two lines of work in RE. One is  distant supervision learning which relies on the  knowledge bases (KBs) to obtain label information \cite{DBLP:conf/tal/ReiplingerWK14:48,DBLP:conf/aaai/Ji0H017:49,jia-etal-2019-arnor:50}. The other is semi-supervised learning ~\cite{Jones_Bootstrapping99,Batista_Bootstrapping15,DBLP:conf/iclr/WangQZ0YN0R20}, which is more general than distant supervision learning in the sense that it does not need the external KB resources. However, most of semi-supervised learning approaches are rule-based ~\cite{Batista_Bootstrapping15,Gupta_Bootstrapping18,DBLP:conf/iclr/WangQZ0YN0R20} and still involve manual operations. More recently, a semi-supervised method DualRE~\cite{DBLP:conf/www/Lin0QR19:13} is proposed to automatically select unlabeled samples to  augment the training set using two collaborative modules. The main advantage of DualRE lies in that it does not need the pre-defined or annotated rules.

Our proposed MRefG model is similar to DualRE in that they are both under semi-supervised learning framework without the usage of rules. The difference is that MRefG exploits the connection between unlabeled instances and labeled ones rather than directly building a mapping function in DualRE. By leveraging the lexical and semantic reference information, our model can capture the joint relations in the data and thus achieves better performance than DualRE.

\paragraph{Graph Neural Networks (GNNs)}
Our work is relevant to graph neural networks  \cite{DBLP:conf/iclr/KipfW17:27,DBLP:conf/nips/HamiltonYL17:28,DBLP:conf/iclr/VelickovicCCRLB18:30}. The target of GNN is to learn for a target node $v$  the representation that contains $v$'s neighborhood information. Several recent attempts in RE have already taken the advantage of the GNN structure. AGGCN~\cite{DBLP:conf/acl/GuoZL19:35} takes full dependency trees as inputs of the graph. GraphRel~\cite{DBLP:conf/acl/FuLM19:41} uses the dependency tree as the sentence's adjacency matrix and uses GCN in addition to LSTM to extract features. A walk-based model \cite{DBLP:conf/acl/ChristopoulouMA18:42} constructs entity-based graphs  to encode the dependencies between relations. All these methods are built upon the intra-sentence graphs where two nodes belong to the same sentence.

To the best of our knowledge, we are the first that exploits the inter-sentence graphs to connect the labeled and unlabeled data. Though the GCNN model~\cite{DBLP:conf/acl/SahuCMA19:46} utilizes both local and nonlocal relations, it is developed for the document-level RE task. More importantly, the inter-sentence edge in ~\cite{DBLP:conf/acl/SahuCMA19:46} represents the co-reference or adjacent sentence relation. In contrast, we build the semantic, verb, and entity edges between two sentences, which are easier to get than the co-reference relation and carry more important information than the adjacent sentence relation.

\section{Proposed Model}
In this section, we first present the problem definition and give the overall structure of the proposed model, and then illustrate each component in the model.
\subsection{Problem Definition and Model Overview}
\textbf{Definition 1} (\emph{Sentence-level Relation Extraction}) Given a sentence $d$ = ($w_1, w_2, ..., w_n$), a subject entity $e_s$ and an object entity $e_o$ in $d$, the task of sentence-level RE is to predict the relation label $r\in{\mathcal{R}}$ between $e_s$ and $e_o$, where $\mathcal{R}$ is a set of pre-defined  relations and the ``no\_relation''.

\noindent\textbf{Definition 2} (\emph{Semi-supervised relation extraction}) Given a set of labeled and unlabeled relation mentions $L=\{(d_i,r_i)\}_{i=1}^{N_L}$ and $U=\{(d_i)\}_{i=1}^{N_U}$, respectively, the goal of semi-supervised RE task is to learn a mapping function that can fit the labeled training data $L$ and capture the information in the unlabeled data $U$ simultaneously.

\begin{figure}[t]
\vspace{-0.2cm}
\center{
\includegraphics[scale=0.4]{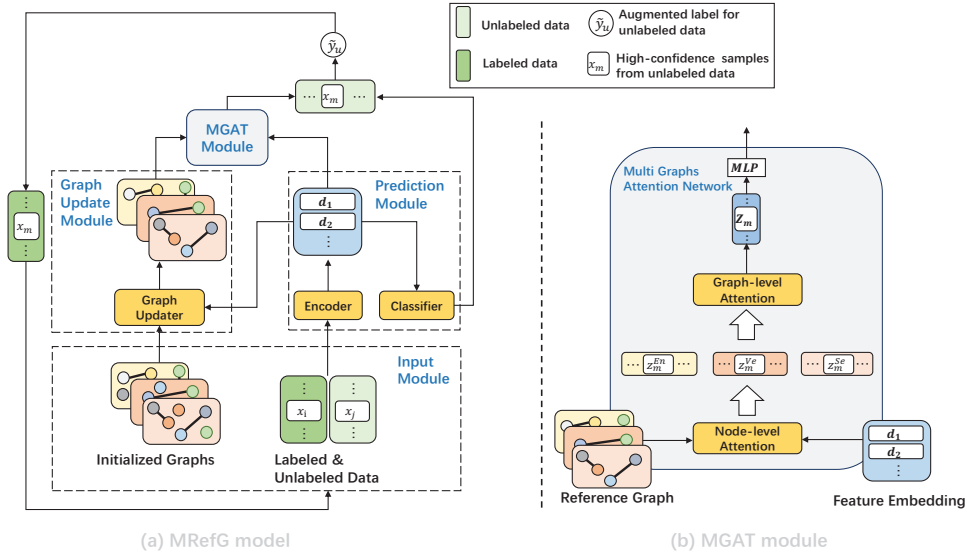}}
\vspace{-0.4cm}
\caption{Model Overview. (a) MRefG model architecture, (b) MGAT module structure.}
\label{fig:architecture}
\vspace{-0.5cm}
\end{figure}

\noindent\textbf{Model Overview}
In this work, we focus on developing a semi-supervised RE method by exploiting the relations between labeled and unlabeled instances.
To this end, we propose a Multiple Reference Graph  (MRefG) model. The overall structure of MRefG is shown in Fig.~\ref{fig:architecture} (a). It consists of 4 modules: (1) an input module to build the initial reference graphs and add the newly identified unlabeled samples into the training set at each iteration, (2) a graph update module to update the semantics reference graph, (3) a MGAT module to leverage multiple reference graphs for identifying high-quality samples from unlabeled data, and (4) a prediction module  to predict labeled and unlabeled samples for model training and data augmentation, respectively.

\subsection{Input Module}
The input module prepares data for other modules. As we illustrate in the introduction, we consider the semantic and lexical relations between labeled and unlabeled samples such that related samples will be placed closely in the vector space. To this end, we first construct three types of graphs: Entity Reference Graph $\mathcal{G}^{e}$, Verb Reference Graph $\mathcal{G}^{v}$, and Semantics Reference Graph $\mathcal{G}^{s}$, where each node denotes a sample sentence and the edge denotes the semantic and lexical relations between two sentences. 
In addition, the input module also merges the newly augmented samples into the training set during the training process, which is a simple operation. Below we give the details of graph construction.

\paragraph{Entity Reference Graph Construction} Entities are critical to RE tasks. To decide whether to add an edge between two sentences by their entity relation, we propose a token level and type level similarity. The former denotes there is an overlap between the entity tokens. For example, given a sentence $d_1$ ``The ride-on \emph{boat tiller} was developed by him'', there are two entities \emph{``boat''} and \emph{``tiller''}. Since two entities in the sentence are adjacent, the other words in the sentence have a weak impact on the entity relation. Now given another sentence $d_2$ ``A \emph{catamaran tiller} having a grip is provided to permit tension'' which also has adjacent entities and an overlapping entity token \emph{``tiller''} with $d_1$, we will add an edge for $d_1$ and $d_2$. The latter denotes that the adjacent entities in two sentences have the same NER (Named Entity Recognition) type set recognized by a parser\footnote{https://stanfordnlp.github.io/CoreNLP/index.html}. For example, two sentences ``Sentenced to death were \emph{Sulta} \emph{Hashim Ahmad al-Tai}'' and ``Voters choosed five republicans to suceed \emph{Rep.} \emph{Paul Gillmor}'' will have an edge, since they both have the NER type set \{PERSON, TITLE\}.


Formally, we define the entity relation edge $e_{ij}$ between two sentences $i$ and $j$ as follows.
\vspace{-0.3cm}
\begin{equation}
\label{ee:construct}
\small
e_{ij}=\begin{cases}
0,\qquad NER_i\neq NER_j \quad and \quad Entity_i \cap Entity_j = \varnothing\\
1,\qquad NER_i=NER_j \quad or \quad Entity_i \cap Entity_j\neq\varnothing
\end{cases}
,
\vspace{-0.2cm}
\end{equation}

where $NER$ is the NER type set, and $Entity$ is the entity token set for two adjacent entities in the sentence, and $i$ is a sample in labeled data, and $j$ is a sample in either labeled or unlabeled data.

\paragraph{Verb Reference Graph Construction} When two entity mentions are not adjacent, the verb or verb phrase (including predicate and non-predicate) between them plays a critical role in the sentence. For example, in ``The \emph{mice} fell into the \emph{trap}'', two entities \emph{`mice}' and \emph{`trap'} are not adjacent. In this case, it is the verb phrase \emph{`fell into'} that determines the subject and the object of the action. In view of this, we construct the verb reference graphs to model the similarity between two samples with the same verb. Formally, we define the verb relation edge $v_{ij}$ between two sentences $i$ and $j$ as follows.
\vspace{-0.3cm}
\begin{equation}
\label{ve:construct}
\small
v_{ij}=\begin{cases}
0, \qquad  V_i \neq V_j\\
1,\qquad   V_i=V_j
\end{cases}
,
\vspace{-0.2cm}
\end{equation}
where $V$ denotes the verb or verb phrase in the sentence, and $i$ is a sample in labeled data, and $j$ is a sample in either labeled or unlabeled data.


\paragraph{Semantics Reference Graph Construction} The semantics denotes the inherent meaning of the sentence beyond the lexical structure. For example, two sentences  ``The \emph{water} was in a \emph{cup}'' and ``The \emph{lawsonite} was contained in a \emph{platinum crucible}'' are not similar from the lexical point of view. However, their meanings are very similar and they should be placed closely in the vector space. We build the semantic edge $s_{ij}$ between two sentences based on their cosine similarity.
\vspace{-0.3cm}
\begin{equation}
\label{se:construct}
\small
s_{ij}=\begin{cases}
0, \qquad   cossim(\bm{d_i},\bm{d_j}) \le \delta\\
1,\qquad    cossim(\bm{d_i},\bm{d_j}) > \delta
\end{cases}
,
\vspace{-0.2cm}
\end{equation}
where $\bm{d}$ is the sentence embedding, and $i$ is a sample in labeled data, and $j$ is a sample in either labeled or unlabeled data. $\delta$ is a predefined threshold which we simply set as 0.9 to get highly related sentences.

\subsection{Prediction Module}
The prediction module is a basic component of semi-supervised models which predicts the labeled and unlabeled samples for model training and data augmentation, and it is also used for predicting the test data after the training is finished. It consists of two parts: an encoder and a classifier. The encoder aims to encode the samples into latent vectors.
There are a number of encoders proposed in the literature such as LSTM \cite{DBLP:journals/neco/HochreiterS97:40}, PCNN \cite{DBLP:conf/emnlp/ZengLC015:21}, and PRNN\cite{DBLP:conf/emnlp/ZhangZCAM17:7}. Since this is not the focus of our research, we simply choose PRNN as the encoder for a fair comparison with the major baseline DualRE~\cite{DBLP:conf/www/Lin0QR19:13}.
PRNN utilizes various types of  text information  as features, including POS (part-of-speech) tags, NER tags, tokens, and entity positions. It then applies a bidirectional LSTM to learn the corresponding representation for each token and  uses the attention mechanism to get the final embedding of the sentence.

As for the classifier, we apply the softmax function to the sentence embedding to get the class label of the relation mentions. Note the predictions on labeled data are used for model training with a cross-entropy loss function:
\vspace{-0.3cm}
\begin{equation}
\small
	\mathcal{L}_\mathcal{P}=-\sum_{l\in\mathcal{Y}_l}y_llog(\Tilde{y}_l^P),
\vspace{-0.2cm}
\end{equation}
where $\mathcal{Y}_l$ is a set of labeled samples, $y_l$ is the true label, and $\Tilde{y}_l^P$ is the  label predicted by the prediction module $P$ for the $l_{th}$ sample in $\mathcal{Y}_l$. On the other hand, the predictions on unlabeled data will be combined with the output from MGAT module to get the final predictions of the unlabeled samples. To clarify, we denote this prediction as $\Tilde{y}^P_u = softmax(\bm d_u)$, where $\bm d_u$ is the  embedding for the sample $u$ in the unlabeled data.

\subsection{Graph Update Module}
In semi-supervised learning, the training data is gradually augmented with the newly labeled samples. At the initial stage, there is a few training samples and it is hard to learn a good embedding for the sentence. Moreover, since the semantic edge is based on the sentence embedding, it might provide wrong reference information with the coarse representation.  We design a graph update module to handle these problems. Firstly, the module will update three graphs by treating the augmented samples as new labeled nodes and adding new edges according to Eqs.~\ref{ee:construct}, ~\ref{ve:construct}, ~\ref{se:construct}. Secondly, the module will remove the edges in the previous semantic graph that are not qualified as a semantic edge under the updated embedding space.


\subsection{Multi Graph Attention Network (MGAT) Module}

The MGAT module is used to exploit multiple reference graphs. It adopts the hierarchical structure and contains a node-level attention layer and a graph-level attention layer.

\paragraph{Node-level Attention Layer} For simplicity, our reference graphs are constructed roughly based on three types of similarity, and the connections may contain many noises. For example, given two sentences ``This week, for example, brought a public \emph{statement}  by the \emph{head} of NOW'', and ``Technological advances have brought the \emph{wars} into the  \emph{living rooms}'', they have the same verb `\emph{brought}' and have an edge between them. However, these two sentences have different relations and the connection information is not consistent with the relation label.
The node-level attention layer is proposed to refine connection information  based on the multi-head self-attention \cite{DBLP:conf/nips/VaswaniSPUJGKP17:43,DBLP:conf/iclr/VelickovicCCRLB18:30}.

Given a sample node pair ($i$, $j$) in the reference graph $\phi$, a single head attention $a_{i,j,k}^{\phi}$ denotes how important the reference node $j$ is to node $i$ in the $k_{th}$ head:
\vspace{-0.3cm}
\begin{equation}
\small
    a_{i,j,k}^{\phi} = MHA(\bm{d_i},\bm{d_j},\phi),
\vspace{-0.2cm}
\end{equation}
where $MHA$ denotes the function defined with a $K$-head attention layer, $\bm{d_i}$ and $\bm{d_j}$ are the feature vectors for node $i$ and node $j$, and $k$ is the head index ($k$ = 1..$K$).
After that, we concatenate the output from each of the $K$-head layers and get the final embedding $\bm z_i^{\phi}$ of node $i$ in the graph $\phi$, i.e.,
\vspace{-0.3cm}
\begin{equation}
\small
	\bm z_i^{\phi} = \mathop{||}\limits_{k=1}^K \sigma{(\sum_{j\in\mathcal{N}_{i}}{\alpha_{i,j,k}^{\phi} \bm W_{k}\bm d_{j}})},
\vspace{-0.2cm}
\end{equation}
where $\bm W_k$ is the linear transformation matrix,  $\alpha_{i,j,k}^{\phi}$ is the  attention coefficient normalized from $a_{i,j,k}^{\phi}$ using a softmax function, $\sigma$ is the sigmoid function, $\mathcal{N}_{i}$ is neighborhood of node $i$. 

\paragraph{Graph-level Attention Layer} Different reference graphs contain various types of information and their contribution to the model varies considerably. Therefore, we add a graph-level attention layer as the second layer in the module to adaptively adjust the weights of different graphs.
Inspired by the HAN model \cite{DBLP:conf/www/WangJSWYCY19:37}, we propose a method that can automatically learn the importance of different graphs to effectively fuse all reference information. Specifically, the weight of the $i_{th}$ node in the graph $\phi$ is calculated as follows:
\vspace{-0.3cm}
\begin{equation}
\small
	w^{\phi}_i=\frac{1}{|\mathcal{V}|}\sum_{i\in\mathcal{V}}\bm{q^T} \cdot tanh(\bm{W}\cdot \bm{z_i}^{\phi}+b),
\vspace{-0.3cm}
\end{equation}
where $\mathcal{V}$ is the node set and $\bm{z_i^\phi}$ is the embedding of the $i_{th}$ node in $\phi$, $\bm{W}$ and $b$ are the weight matrix and bias, $\bm{q}$ is the graph-level attention vector randomly initialized and updated in the network.

We then use a softmax function to normalize the weight $w^{\phi}_i$.
\vspace{-0.3cm}
\begin{equation}
\small
	\beta^{\phi}_i=softmax(w^{\phi}_i)=\frac{exp(w^{\phi}_i)}{\sum_{\phi=1}^M exp(w^{\phi}_i)},
\vspace{-0.2cm}
\end{equation}
where $M$ (=3) is the number of reference graphs.  With the normalized graph-level importance weights, we aggregate all the graph-specific embeddings to get the final representation $\bm z_i$ for node $i$.
\vspace{-0.3cm}
\begin{equation}
\small
	\bm{z_i}=\sum_{\phi=1}^M{\beta^{\phi}_i\cdot \bm{z^{\phi}_i}},
\vspace{-0.2cm}
\end{equation}

Since each node $i$ has a graph-specific embedding $\bm z_i$ from the MGAT module, we can feed $\bm z_i$ into a one-layer MLP to get the prediction  $\Tilde{y_i}$ of the relation mention in $i$. Once again, for the labeled samples, we utilize cross-entropy as the loss function to train the model by minimizing it.
\vspace{-0.3cm}
\begin{equation}
\small
	\mathcal{L}_\mathcal{M}=-\sum_{l\in \mathcal{Y}_l} y_l log(\Tilde{y}_l^M),
\vspace{-0.3cm}
\end{equation}
where $\mathcal{Y}_l$ is the labeled node set, $y_l$ is the true label, $\Tilde{y}_l^M$ is the  label predicted by the MGAT module.

The entire model is trained by alternatively optimizing $\mathcal{L}_\mathcal{P}$ and $\mathcal{L}_\mathcal{M}$.
The prediction for an unlabeled sample by this module is denoted as $\Tilde{y}^M_u$, which is used to intersect with $\Tilde{y}^P_u$ generated by the prediction module to get the final label of the unlabeled sample $d_u$, and the top 10\% percent of unlabeled samples will be chosen for augmenting the  labeled data.


\section{Experiment}
\subsection{Dataset}
We use two public datasets in our experiments: SemEval and TACRED.

(1) \textbf{SemEval} (SemEval-2010 Task8) \cite{DBLP:conf/semeval/HendrickxKKNSPP10:39} is the most commonly used dataset for RE tasks. It contains 18 relations like ``\emph{Product-Producer (e2,e1)}'' and an extra ``\emph{no\_relation}''.

(2) \textbf{TACRED} (TAC Relation Extraction Dataset) \cite{DBLP:conf/emnlp/ZhangZCAM17:7} is a large-scale RE dataset which consists of news lines and online texts and is used in the annual TAC knowledge base population competition. It includes 41 relation types such as ``\emph{per:age}'' and an extra ``\emph{no\_relation}''.

Note that the relations in SemEval are directional, i.e., $RE(e_1,e_2) \neq RE(e_2,e_1)$. The relations in TACRED do not consider direction. The detailed descriptions of datasets are shown in Table \ref{dataset}.

\begin{table}[h]
\vspace{-0.4cm}
\caption{\label{dataset} Statistics for datasets. }
\setlength{\tabcolsep}{1mm}
\small
\begin{center}
\begin{tabular}{cccccc}
\hline
\textbf{Dataset} & \textbf{\#Train} & \textbf{\#Dev} & \textbf{\#Test} & \textbf{\#Relations} & \textbf{\%No\_relation}\\
\hline
SemEval  & 7,199 & 800 & 2,715 & 19 & 17.6\\
TACRED  & 68,124 & 22,631 & 15,509 & 42 & 79.5\\
\hline
\end{tabular}
\end{center}
\vspace{-0.3cm}
\end{table}

We use the data split provided by DualRE~\cite{DBLP:conf/www/Lin0QR19:13}, i.e., 5\%, 10\%, 30\% of training data from SemEval, and 3\%, 10\%, 15\% from TACRED are used as labeled sets, and 50\% of training samples are used as unlabelled data for both datasets.

\subsection{Compared Methods and Settings}
We compared our MRefG model with the following 8 baselines and 1 method using gold labels.

(1) \textbf{LSTM} \cite{DBLP:journals/neco/HochreiterS97:40}  is a widely used method in text classification tasks.

(2) \textbf{AGGCN} \cite{DBLP:conf/acl/GuoZL19:35} utilizes GCN with multi-head attention to extract important information from the dependency tree. Note that AGGCN does not distinguish the direction of the relation since it discards the position of entity words.

\begin{table}[t]
\tiny
\vspace{-0.0cm}
\caption{\label{SemEval} Comparison results on SemEval. The best and the second best scores are in bold and underlined.}
\vspace{-0.2cm}
\setlength{\tabcolsep}{1mm}
\begin{center}
\begin{adjustbox}{center}
\begin{tabular}{lccccccccc}
\hline
Methods/LabeledData && 5\% &&& 10\% &&& 30\%& \\
 & Precision & Recall & $F_1$ & Precision & Recall & $F_1$ & Precision & Recall & $F_1$ \\
\hline
LSTM & $24.26\pm1.76$ & $21.04\pm1.63$ & $22.52\pm1.62$ & $40.14\pm5.08$ & $33.08\pm4.01$ & $35.88\pm2.17$ & $64.13\pm2.46$ & $63.78\pm3.03$ & $63.84\pm0.55$\\
AGGCN & $8.26\pm4.44$ & $5.07\pm0.49$ & $5.99\pm1.26$ & $9.70\pm3.51$ & $4.83\pm0.30$ & $6.34\pm0.86$ & $8.69\pm1.93$ & $5.32\pm0.25$ & $6.60\pm0.47$\\
PCNN & $41.56\pm2.51$ & $39.30\pm3.56$ & $40.30\pm2.49$ & $53.68\pm1.26$ & $49.87\pm1.50$ & $51.66\pm1.38$ & $64.49\pm0.64$ & $62.81\pm0.55$ & $63.37\pm0.42$\\
PRNN & $55.65\pm1.34$ & $53.73\pm1.25$ & $54.66\pm0.89$ & $63.47\pm3.14$ & $61.76\pm2.20$ & $62.49\pm0.59$ & $69.66\pm2.19$ & $68.76\pm2.60$ & $69.14\pm1.02$\\
\hline
NERO  & $68.00\pm1.75$ & $50.631\pm1.06$ & $58.01\pm0.19$ & $66.40\pm0.81$ & $51.96\pm1.23$ & $58.28\pm0.52$ & $70.12\pm0.26$ & $52.28\pm0.61$ & $59.90\pm0.27$\\
Self-Training  &  $60.60\pm2.19$ & $55.59\pm1.21$ & $57.95\pm0.83$ & $63.92\pm0.48$ & $67.28\pm0.82$ & $65.55\pm0.30$ & $70.33\pm1.08$ & $73.56\pm1.12$&$71.89\pm0.19$\\
RE-Ensemble  & $58.78\pm1.41$ & $59.21\pm1.46$ & $58.97\pm0.76$ & $63.63\pm0.74$ & $67.03\pm0.64$ & $65.29\pm0.27$ & $70.51\pm1.03$ & $73.51\pm1.28$ & $71.97\pm0.43$\\
DualRE & $60.10\pm0.94$ & $61.94\pm1.55$ & $\underline{61.10\pm1.11}$  & $64.14\pm0.83$ & $68.32\pm0.68$ & $\underline{66.16\pm0.43}$ & $70.56\pm0.76$ & $74.21\pm0.33$ & $\underline{72.33\pm0.32}$\\
MRefG (ours)  & $60.02\pm0.52$ & $64.57\pm1.07$ &  $\bm{62.21}\pm\bm{0.48}$ & $65.00\pm0.59$ & $69.58\pm0.26$ & $\bm{67.17}\pm\bm{0.35}$ & $70.97\pm0.78$ & $74.86\pm0.62$ & $\bm{72.86}\pm\bm{0.44}$\\
\hline
RE-Gold(PRNN) & $72.25\pm0.97$ & $75.54\pm1.98$ & $73.83\pm0.62$ & $71.40\pm1.42$ & $76.72\pm0.64$ & $73.95\pm0.50$ & $72.98\pm0.96$ & $78.86\pm0.76$ & $75.80\pm0.24$\\
\hline
\end{tabular}
\end{adjustbox}
\end{center}
\vspace{-0.3cm}
\end{table}

\begin{table}[htp]
\vspace{-0.3cm}
\tiny
\caption{\label{TACRED} Comparison results on TACRED. The best and the second best scores are in bold and underlined.}
\vspace{-0.1cm}
\setlength{\tabcolsep}{1mm}
\begin{center}
\begin{adjustbox}{center}
\begin{tabular}{lccccccccc}
\hline
Methods/Labeled Data && 3\% &&& 10\% &&& 15\%& \\
 & Precision & Recall & $F_1$ & Precision & Recall & $F_1$ & Precision & Recall & $F_1$ \\
\hline
LSTM & $38.76\pm3.19$ & $18.43\pm3.78$ & $24.55\pm2.79$ & $48.72\pm1.57$ & $44.55\pm1.63$ & $46.82\pm0.63$ & $53.83\pm0.73$ & $46.79\pm2.11$ & $50.02\pm0.94$\\
AGGCN & $51.95\pm5.16$ & $38.65\pm5.51$ & $43.69\pm1.34$ & $59.80\pm1.00$ & $51.77\pm0.95$ & $55.48\pm0.39$ & $61.30\pm0.82$ & $52.69\pm0.30$ & $56.66\pm0.18$\\
PCNN & $53.89\pm3.29$ & $39.93\pm2.47$ & $45.39\pm0.78$ & $64.66\pm3.16$ & $41.84\pm2.63$ & $50.42\pm1.00$ & $66.85\pm0.43$ & $42.90\pm1.06$ & $52.25\pm0.67$\\
PRNN & $42.85\pm1.09$ & $39.52\pm1.65$ & $41.07\pm0.51$ & $53.44\pm2.82$ & $51.77\pm1.88$ & $52.49\pm0.64$ & $58.88\pm2.32$ & $51.30\pm2.04$ & $54.76\pm0.93$\\
\hline
NERO  & $47.89\pm0.58$ & $44.21\pm0.45$ & $45.97\pm0.41$ & $47.2\pm0.93$ & $46.94\pm0.79$ & $47.07\pm0.61$ & $51.57\pm1.40$ & $43.98\pm1.91$ & $47.48\pm0.92$\\
Self-Training  & $53.27\pm1.08$ & $38.53\pm1.64$ & $44.70\pm1.38$ & $57.19\pm1.10$ & $53.73\pm1.09$ & $53.39\pm0.15$ & $59.92\pm0.64$ & $55.71\pm1.48$ & $57.72\pm0.64$\\
RE-Ensembling  &  $54.68\pm2.99$ & $39.93\pm1.41$ & $46.07\pm0.44$ & $59.04\pm1.27$ & $53.06\pm0.97$ & $55.87\pm0.30$ & $60.77\pm1.11$ & $55.57\pm0.87$ & $57.81\pm0.30$\\
DualRE  & $52.60\pm0.82$ & $41.64\pm1.06$ & $\underline{46.47\pm0.27}$ & $58.96\pm1.33$ & $54.64\pm0.82$ & $\underline{56.69\pm0.31}$ & $59.74\pm1.24$ & $56.20\pm0.30$ & $\underline{57.92\pm0.23}$\\
MRefG (ours) & $50.35\pm1.55$ & $47.27\pm3.75$ &  $\bm{46.91}\pm\bm{0.86}$ & $59.90\pm1.01$ & $51.77\pm0.95$ & $\bm{57.17}\pm\bm{0.24}$ & $56.96 \pm 0.59$ & $59.80\pm0,48$  & $\bm{58.34}\pm\bm{0.28}$\\
\hline
RE-Gold(PRNN)  & $65.63\pm1.68$ & $57.47\pm1.18$ & $61.25\pm0.04$ & $65.06\pm4.26$ & $60.18\pm3.16$ & $62.31\pm0.27$ & $66.13\pm1.82$ & $59.62\pm2.35$ & $62.64\pm0.49$\\
\hline
\end{tabular}
\end{adjustbox}
\end{center}
\vspace{-0.6cm}
\end{table}


(3) \textbf{PCNN} \cite{DBLP:conf/emnlp/ZengLC015:21} employs a piecewise CNN with position embedding to model word sequences.

(4) \textbf{PRNN} \cite{DBLP:conf/emnlp/ZhangZCAM17:7} utilizes bi-LSTM  to encode position, POS, NER, and token information. It is the state-of-the-art supervised method for RE tasks.


(5) \textbf{NERO} \cite{DBLP:conf/www/ZhouLLWDNR20} is a rule-based method. It first obtains patterns with manual annotation from the data, and then uses soft-matching for extracting relations. For a fair comparison, we use both labeled and unlabelled data as training set for rule matching in NERO.

(6) \textbf{Self-Training} \cite{DBLP:conf/wacv/RosenbergHS05:24} is a traditional semi-supervised method. It utilizes the prediction information on  unlabeled data and adds the top-ranked samples into the training set.

(7) \textbf{Re-Ensemble} \cite{DBLP:conf/iclr/FrenchMF18:23} constructs two independent prediction modules and then uses the intersection predicted by these two modules to get high-confidence label information.

(8) \textbf{DualRE} \cite{DBLP:conf/www/Lin0QR19:13} jointly trains a prediction and a retrieval module to select samples from unlabeled data. We use the point-wise variant as it performs better than the other pairwise variant.

(9) \textbf{RE-Gold} uses PRNN as the base model but treats both labeled and unlabeled corpus as the labeled data and thus it can provide the upper bound for semi-supervised methods.

Among the baselines, the first four methods and RE-Gold are supervised and the remains are semi-supervised ones.
For the baselines with released code, we re-train them using the optimal hyper-parameters reported in their original papers. For those without source code, we re-implement and re-train them under the same settings as our model. Specifically, we use 300-dimension GloVe \cite{DBLP:conf/emnlp/PenningtonSM14:44} as pre-trained word embeddings and use Stanford CoreNLP tool \cite{DBLP:conf/acl/ManningSBFBM14:45} to extract dependency information. Other parameters unique to our model are adjusted by the development set.

Moreover, following DualRE, we choose PRNN as the encoder to encode additional information for all semi-supervised methods. The only exception is NERO \cite{DBLP:conf/www/ZhouLLWDNR20} which uses a bi-LSTM as encoder since the position and NER  information has already included during preprocessing.

\vspace{-0.1cm}
\subsection{Main Results}
\vspace{-0.1cm}
The comparison results on SemEval and TACRED are shown in Table~\ref{SemEval} and Table~\ref{TACRED}, respectively.
We divide the results into three parts. The first and the second part are supervised  and semi-supervised methods, while the third one is the upper bound method. We have the following important findings.

(1) It is clear that our MRefG model is the best among all methods.  Also, the gap between MRefG and the upper bound RE-Gold method becomes narrow with the increasing number of labeled data.

(2) Among the supervised baselines, PRNN and AGGCN performs the best on SemEval and TACRED, respectively.  The extremely poor performance of AGGCN  on SemEval can be due to that it is not designed for direction-sensitive RE tasks.

(3) Semi-supervised methods outperform supervised ones.  DualRE is the best semi-supervised baseline owing to the carefully designed retrieval model. However, it is still inferior to our model. This proves that our model benefits from the reference graphs by mapping the similar samples to the same relation. In contrast to the results in its original paper \cite{DBLP:conf/www/ZhouLLWDNR20}, NERO performs worse than DualRE in our experiment. The reason might be twofold. Firstly, the NERO paper uses all training data while we use at most (30+50)\% and (15+50)\% training data on SemEval and TACRED. Note this is necessary for examining semi-supervised learning methods, and it is fair since all methods are under the same setting. Secondly, the NERO paper adopts bi-LSTM and attention as the encoder for DualRE without considering POS, NER, and position, thus the important information is missing from DualRE.

\section{Analysis}
\subsection{Ablation Study}
To investigate the impacts of different reference graphs in our MRefG model, we conduct the ablation study by using one graph at each time.  The results on two datasets are shown  in Table \ref{Ablation}, where ``Original'' denotes the results for the original  model, and ``Verb'', ``Entity'', and ``Sematic'' denotes the results of keeping the corresponding graph and removing two others from the original model.
\vspace{-0.4cm}
\begin{table}[h]
\caption{\label{Ablation} Ablation study for MRefG. $\downarrow$ denotes the drop of F1 score.}
\vspace{-0.2cm}
\setlength{\tabcolsep}{1mm}
\small
\begin{center}
\begin{tabular}{ccccccc}
\hline
& \multicolumn{3}{c}{10\%SemEval} & \multicolumn{3}{c}{10\%TACRED} \\ \cline{2-7}
 & precision & recall & F1 & precision & recall & F1 \\ \hline
Original & 65.65 & 69.48 & 67.51 & 60.17 & 55.06 & 57.50 \\
Verb & 64.32 & 68.42 & 66.31$\downarrow$ & 57.25 & 55.07 & 56.14$\downarrow$ \\
Entity & 61.59 & 68.02 & 64.65$\downarrow$ & 55.12 & 58.26 & 56.66$\downarrow$ \\
Semantics & 62.38 & 66.87 & 64.55$\downarrow$ & 57.02 & 54.74 & 55.85$\downarrow$ \\ \hline
\end{tabular}
\end{center}
\vspace{-0.5cm}
\end{table}

As expected, the performance of all simplified variants drops, showing that each reference graph contributes to our model. In addition, the single semantics reference graph performs the worst. The reason might be that it relies on the sentence embedding to compute the semantic similarity. With the limited number of labeled data, it is hard to obtain the precise representation.

\subsection{Parameter Sensitivity Analysis}
\begin{figure}[h]
\center{
\includegraphics[scale=0.24]{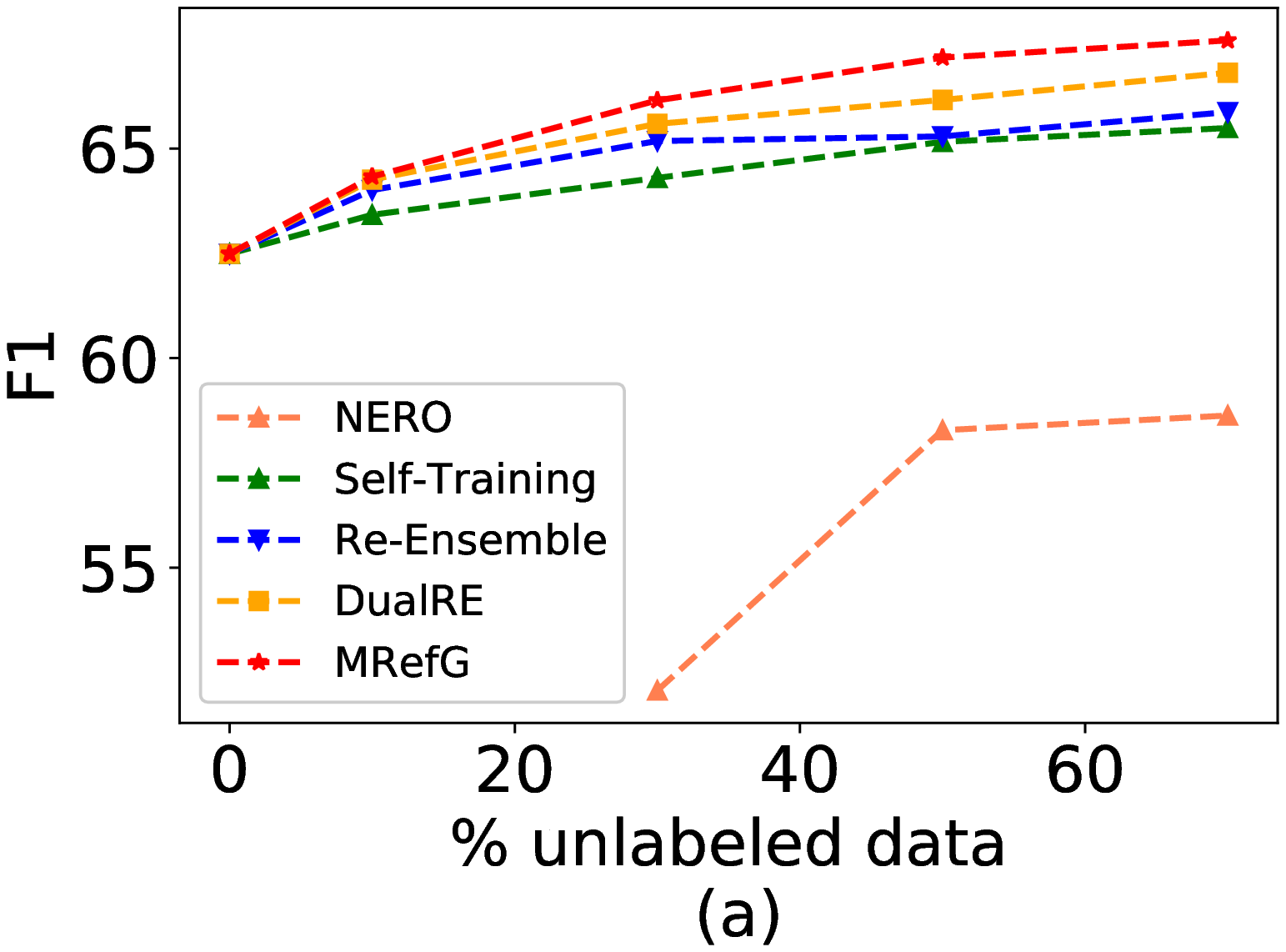}
\includegraphics[scale=0.24]{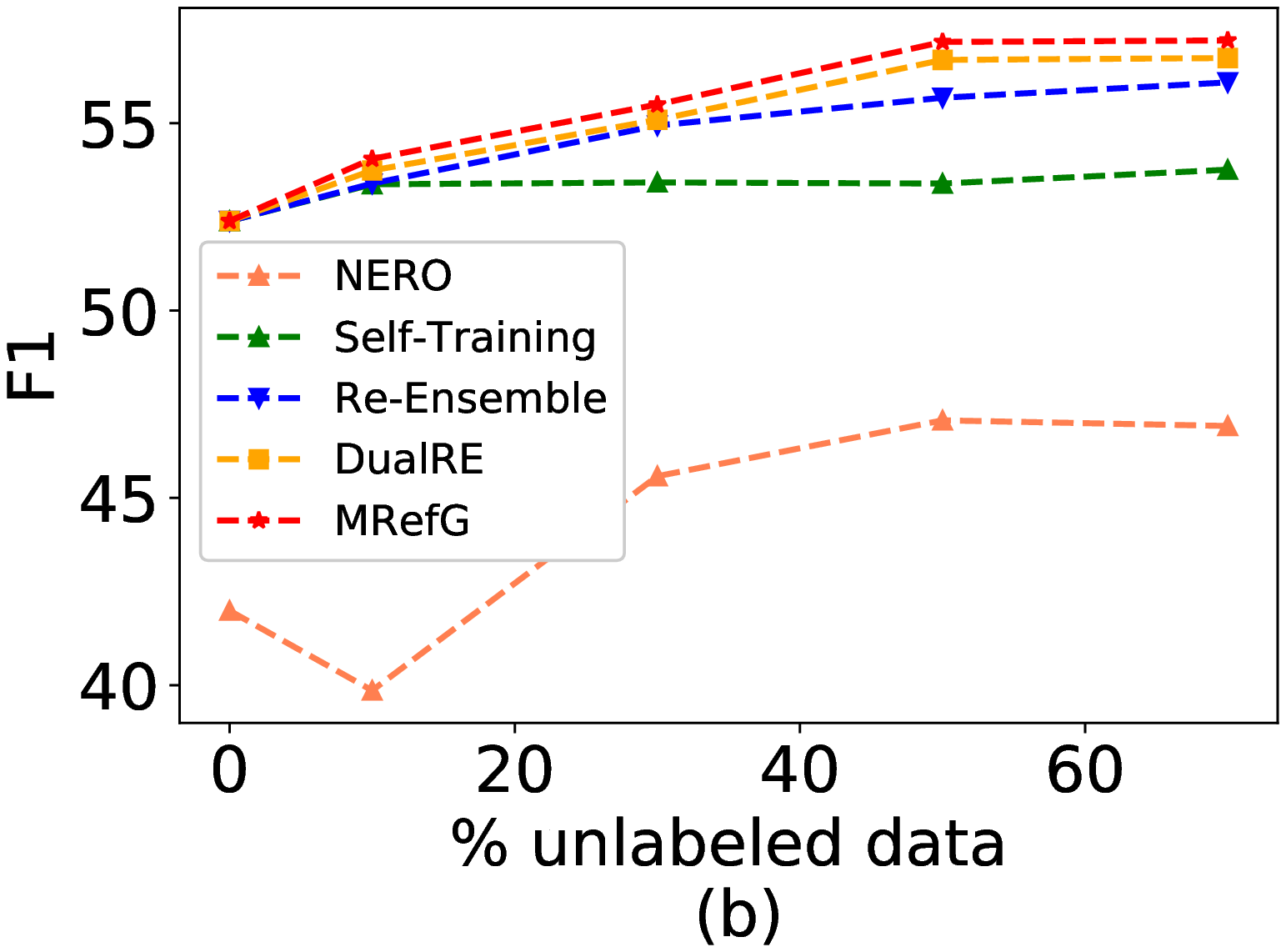}
\includegraphics[scale=0.24]{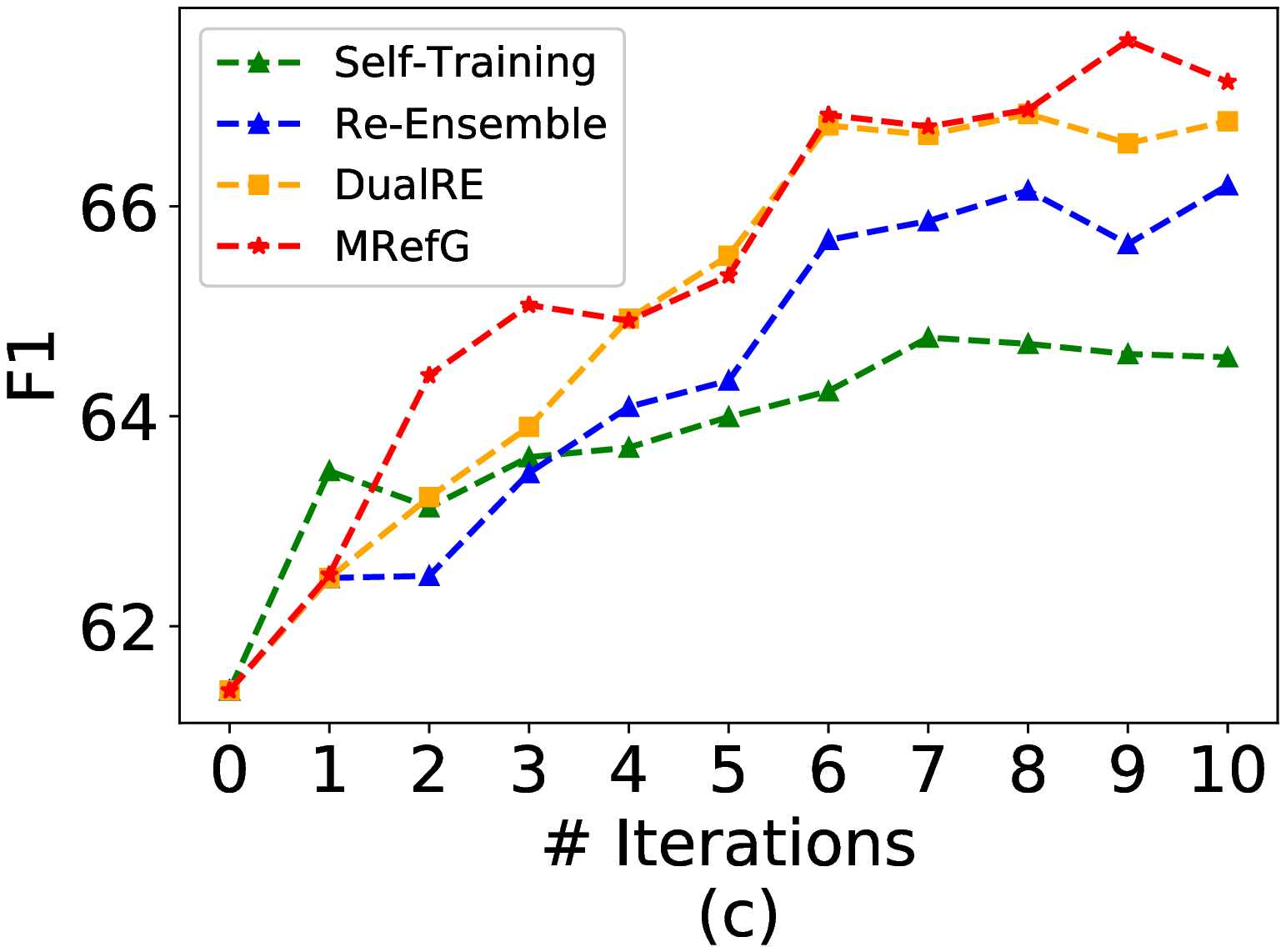}
\includegraphics[scale=0.24]{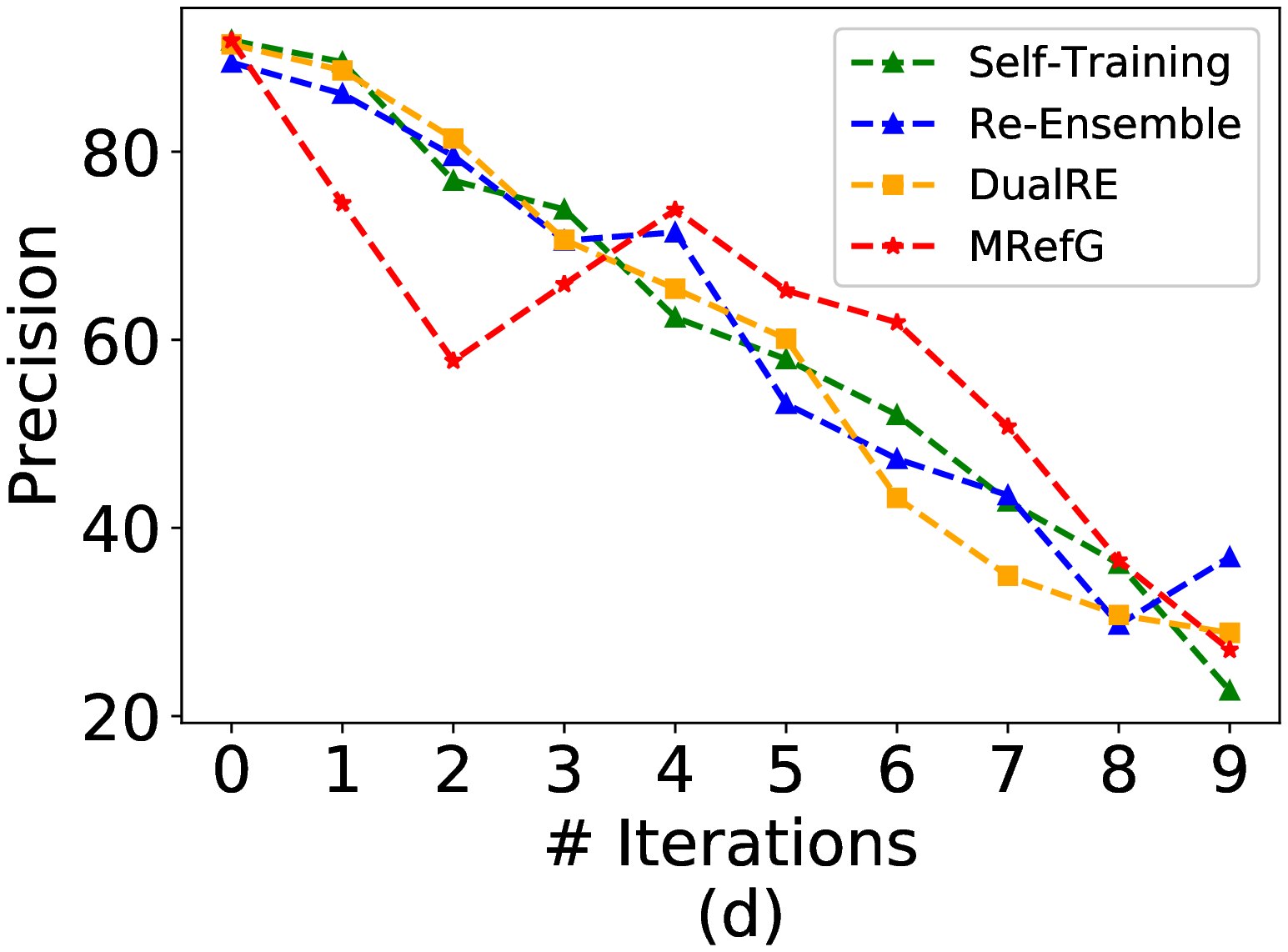}}
\vspace{-3mm}
\caption{Results for detailed analysis. (a) Impacts of the ratio of unlabeled data in SemEval, (b) Impacts of the ratio of unlabeled data in TACRED, (c) Convergence curve of test F1 on SemEval, (d) Precision of the augmented samples on SemEval.}
\label{fig:percent}
\vspace{-3mm}
\end{figure}

This section we study the parameter sensitivity of our model.
We first examine the influence of the size of unlabeled data on semi-supervised methods. We vary the unlabeled data size to 0\%, 10\%, 30\%, 50\%, 70\% of the training set and keep the labeled data to 10\% for both datasets. The results are shown in Fig.~\ref{fig:percent} (a) (b).
As can be seen, our model achieves the best results under various settings. This further proves the effectiveness of our MRefG model on utilizing the reference information. Moreover, with the increasing ratio of unlabeled data size, the difference between our model and others becomes more clear. This implies that our model can better leverage the unlabeled data. Also note that the F1 score of NERO on SemEval is extremely poor when  0\% and 10\% training data is used as unlabeled set, i.e., its F1 is 12.74 and 12.87 for 0\% and 10\%, respectively.
For clearly presenting the curves for other methods, we omit these two results for NERO.

We then investigate the quality of augmented samples on SemEval. Fig.~\ref{fig:percent} (c) (d) visualizes the test F1 score and precision of the augmented samples under different number of iterations using 10\% and 50\% labeled  and unlabeled data.  Due to the space limitation, we only show the results on SemEval. Also note NERO does not has an iteration process, thus it is omitted in Fig.~\ref{fig:percent} (c) (d).



As shown in Fig.~\ref{fig:percent} (c), at the early stage of training, the performance curve of all models has an obvious upward trend, showing that semi-supervised learning benefits from the newly added samples. When the training proceeds, the curve of Self-Training becomes stable, while our MRefG and other two models can still get improvements.
In Fig.~\ref{fig:percent} (d), the samples selected by our model are not as good as others at the early stage since the reference graphs contain lots of noises at that time. However, in the middle and late stage of training, the graphs are gradually refined and enlarged. Our model obtains more reference information through reference graphs, which helps our model to select the high-quality samples from the unlabeled data.  As a result, the precision of new selected samples of our model becomes higher than other models.

\subsection{Case Study}
To have a close look at the impacts of different reference graphs in our model, we perform a case study by comparing it with the strongest baseline DualRE on 4 sentences (abbreviated as S1...S4). The results are shown in Table~\ref{casestudy}.

\vspace{-0.2cm}
\begin{table}[ht]
\small
\caption{\label{casestudy} Case study. The red and blue marked token denotes subject ($e_1$) and object ($e_2$) entity, and \tiny{\CheckmarkBold} \normalsize{and} \tiny{\XSolidBrush} \normalsize{denotes a correct  and wrong prediction of the relation between two entities, respectively}.}
\vspace{-0.0cm}
\setlength{\tabcolsep}{1mm}
\begin{center}
\begin{tabular}{lll}
\hline
\bf Sentence  & \bf DualRE & \bf MRefG \\
\hline
S1: A test \textcolor{red}{payment} was put into my bank \textcolor{blue}{account} \\on October 13th.  & Entity-Destination (e1, e2) (\tiny{\CheckmarkBold}) & Entity-Destination (e1, e2) (\tiny{\CheckmarkBold})\\
\hline
S2: An \textcolor{red}{album} by the same artist has now \\been put into a \textcolor{blue}{folder}.  & No\_Relation (\tiny{\XSolidBrush}) & Entity-Destination (e1, e2) (\tiny{\CheckmarkBold}) \\
\hline
S3: Now the state development bank plans to finance \\the export of ethanol knowhow to Africa, where the \\hopes of countries are pinned on the \textcolor{red}{jatropha} \textcolor{blue}{plant}. & No\_relation (\tiny{\XSolidBrush}) & Product-Producer (e1, e2)  (\tiny{\CheckmarkBold}) \\
\hline
S4: The kind of treatments necessary to kill \\the  \textcolor{red}{bacteria} which cause acne \textcolor{blue}{breakouts} \\must be prescribed by a doctor.  & Cause-Effect (e2, e1) (\tiny{\XSolidBrush}) & Cause-Effect (e1, e2) (\tiny{\CheckmarkBold})\\
\hline
\end{tabular}
\end{center}
\vspace{-0.3cm}
\end{table}

S1 is simple and has a clear structure. Both our model and DualRE can correctly classify the relation.

In S2, the subject entity ``\emph{album}'' has a long modifier ``\emph{by the same artist}'', and DualRE is unable to identify the relation between two entities. In contrast, by building the verb edge between S2 and other training samples containing ``\emph{put into}'',  our model can successfully recognize the ``\emph{Entity-Destination}'' relation for two entities ``\emph{album}'' and ``\emph{folder}''.



S3 has a complicated structure and DualRE is easily affected by other components in the sentence and cannot identify the important features. Consequently, it makes a wrong prediction for the entity pair (``\emph{jatropha}'',``\emph{plant}''). Our model, on the other hand, can  assign the  correct ``\emph{Product-Producer}'' relation to  the entity pair. Through a deep analysis, we find that S3 has  several entity edges connected with the samples in the training set. Here we give two examples. One is ``It is the last \emph{auto} \emph{plant} operating in California''. The other is ``Eighty years on from the first pipes being laid for the \emph{car} \emph{plant}, unemployment is at 2 million and rising''. These two training sentences both have two adjacent entities and a relation label ``\emph{Product-Producer}''. Clearly, it is the entity edge that helps our MRefG model to make correct prediction.



S4 also has a complex structure. Since most of training samples containing ``\emph{cause}'' are in the form of ``\emph{e1}...caused by...\emph{e2}'', their relations are ``\emph{Cause-Effect (e2, e1)}'', which misleads DualRE about the direction of the relation. In our case, S4 has a semantic edge with a training sample ``By avoiding or limiting contact with trash the chance of contracting \emph{germs}  that lead to \emph{sickness} or disease is decreased''.
This sentence has a similar meaning and structure with S4. Moreover, the verb ``lead to'' in its attributive clause is the synonym of ``cause''. Both these provide our model with the helpful hints to give the correct direction of the relation.

\section{Conclusion}
In this paper, we propose a Multiple Reference Graph (MRefG) model for semi-supervised relation extraction. The main idea is to provide the reference information for the unlabeled samples before they are sent into the classifier. To this end, we construct three types of reference graphs to connect labeled and unlabeled samples to make the similar instances positioned closely in the encoded vectors space.  As a result, the mapping function can better predict the relation label for the referees. By choosing such high-confidence samples from the unlabeled corpus, the training set is successfully augmented during each iteration of semi-supervised learning process. The extensive experimental results on two public datasets demonstrate that our proposed model consistently outperforms the state-of-the-art baselines.

\clearpage
\bibliographystyle{coling}
\bibliography{coling2020}

\end{document}